\documentclass{INTERSPEECH2023}

\usepackage{hyperref}
\usepackage{todonotes}
\usepackage{xspace}
\usepackage{multirow}
\usepackage{ amssymb }
\usepackage{scrextend}

\newcommand{\repo}{\url{https://github.com/RiTA-nlp/ITALIC}}

\newcommand{\poli}{$^{\diamondsuit}$}
\newcommand{\enna}{$^{\clubsuit}$}
\newcommand{\bocconi}{$^{\heartsuit}$}

\interspeechcameraready

\title{ITALIC: An ITALian Intent Classification Dataset}
\title{ITALIC: An Italian Intent Classification Dataset}

\name{Alkis Koudounas\poli, Moreno La Quatra\enna, Lorenzo Vaiani\poli, Luca Colomba\poli, Giuseppe Attanasio\bocconi, Eliana Pastor\poli, Luca Cagliero\poli, Elena Baralis\poli}
\address{
  \poli Politecnico di Torino, Turin, Italy\\
  \enna Kore University of Enna, Enna, Italy \\
  \bocconi Bocconi University, Milan, Italy}
\email{\{name.surname\}@polito.it, moreno.laquatra@unikore.it, giuseppe.attanasio3@unibocconi.it}

\begin{document}

\maketitle
 
\begin{abstract}

\noindent
Recent large-scale Spoken Language Understanding datasets focus predominantly on English  
and do not account for language-specific phenomena such as particular phonemes or words in different lects.
We introduce \textsc{ITALIC}, the first large-scale speech dataset designed for intent classification in Italian.
The dataset comprises 16,521 crowdsourced audio samples recorded by 70 speakers from various Italian regions and annotated with intent labels and additional metadata.
We explore the versatility of \textsc{ITALIC} by evaluating current state-of-the-art speech and text models. Results on intent classification suggest that increasing scale and running language adaptation yield better speech models, monolingual text models outscore multilingual ones, and that speech recognition on \textsc{ITALIC} is more challenging than on existing Italian benchmarks.
We release both the dataset and the annotation scheme to streamline the development of new Italian SLU models and language-specific datasets.

\end{abstract}

\section{Introduction}


Spoken Language Understanding (SLU) is pivotal in enabling human-machine interaction through natural language. However, language distribution in high-quality SLU resources is skewed toward a small set of languages, particularly English \cite{LugoschRITB19,BastianelliVSR20}. 
Prior work proposed training resources for languages other than English to let models learn language-specific phonemes, words, or phrases; however, such resources are either not specifically designed for human-machine interaction \cite{common_voice_ds} or lack audio recordings in the target language, preventing end-to-end (E2E) learning from speech \cite{Almawave-SLU,massive_dataset}.  
   
In this paper, we introduce \textsc{ITALIC}, the first large-scale ITAlian Language Intent Classification audio dataset, including 16,521 audio samples spanning 18 domains, 60 intents, and recorded by 70 speakers from a variety of Italian regions. 
To build the collection, we extracted and annotated all Italian interactions in the MASSIVE dataset~\cite{massive_dataset} 
and enriched them by annotating every recording with speaker- and channel-related attributes. We provide the speaker's self-declared region of origin, gender, age, and instruction level, the level of background noise, and the type of recording device used.

This metadata allows a variety of additional analyses beyond intent classification, such as speaker recognition, 
text-to-speech, age estimation, and linguistic variety identification.

To highlight the versatility of \textsc{ITALIC} and the richness of the provided metadata, we benchmark current state-of-the-art speech and text models on the intent classification and automatic speech recognition tasks. The experimental results show that 1) model scale and ASR adaptation improve the performance of speech models in terms of generalization to unseen speakers and robustness to noise, 2) monolingual text models outperform multilingual ones, and 3) zero-shot speech recognition performs worse than existing Italian benchmarks.

\noindent Our contributions are as follows:

\begin{itemize}
    \item We introduce the first Italian large-scale intent classification dataset with recordings and transcripts of virtual assistant utterances. We enrich every instance with self-declared information about the speaker and the recording channel.
    \item We provide several baselines with current state-of-the-art speech and text models. Through experimental results, we show the strengths and weaknesses of such models and highlight the most promising direction to improve SLU models.
    \item We release the dataset, annotation scheme, and code for the baselines to encourage further research on the collection and, more broadly, on SLU in Italian. 

\end{itemize}

\section{Related Work}

Most well-known SLU benchmark datasets are mainly in English. For example, Fluent Speech Commands~\cite{LugoschRITB19} is an open-source SLU dataset including
31 intents and 30,043 English utterances. 
SLURP~\cite{BastianelliVSR20} is 
also an English-only dataset with single turn user interactions with a home assistant.
MASSIVE~\cite{massive_dataset} extends the latter by including more than one million utterances across 51 languages with annotations for Natural Language Understanding (NLU) tasks, but no non-English audio recordings. 
AUDIO SNIPS~\cite{AudioSnips} is the audio version of the SNIPS NLU dataset. It contains both audio samples in English and French annotated with the corresponding intents.

Limited efforts have been made to develop speech and language understanding systems specifically for the Italian language. 
One such attempt is the AlmaWave-SLU~\cite{Almawave-SLU}, which involves the generation of an Italian data collection derived from English AUDIO SNIPS utterances through speech transcription and machine translation. 
However, this corpus lacks Italian audio recordings, rendering it unsuitable for Italian SLU tasks.
Mozilla Common Voice~\cite{common_voice_ds} and Google Fleurs~\cite{fleurs_ds} include Italian audio recordings with transcriptions. However, both datasets do not provide intent annotations nor metadata on the recording conditions or the speakers’ regional origins.
EMOVO~\cite{costantini-etal-2014-emovo} collects emotional speech recordings from six native Italian speakers.
IDEA~\cite{IDEA} is a dataset for modeling dysarthric speech and includes isolated words, recorded under controlled conditions, covering a wide range of phonemes. 
Differently from \textsc{ITALIC}, these Italian datasets are not explicitly created for SLU tasks.

\begin{table}[!h]
    \caption{Gender and age distribution in ITALIC.}
    \label{tab-data-demo}
    \centering
    \begin{tabular}{cc|cccc}
    \toprule
    \multicolumn{2}{c|}{Gender} & \multicolumn{4}{c}{Age} \\ 
    Female & Male & [18-25] & [26-40] & [41-55]  & $\ge$56   \\  \midrule
    42.96\% & 57.04\% & 10.63\% & 63.86\% & 10.78\% & 14.73\% \\ \bottomrule
    \end{tabular}
\end{table}

\section{The ITALIC Dataset}

\subsection{Data collection}
The \textsc{ITALIC} dataset was crowdsourced through a custom web platform. Both native and non-native Italian speakers participated. 
We required participants to record themselves while reading a short instruction randomly sampled from the MASSIVE~\cite{massive_dataset} dataset.
The latter consists of utterance transcripts and associated intents. We use the transcripts as prompts for crowd workers to read out and record. 
We do not annotate intents and instead use those supplied by MASSIVE.
After giving the participants a list of annotation guidelines, we did not intervene in the process or supervise the recording sessions.
They used their own devices and chose freely when and how to contribute recordings.

Optionally, participants could declare additional information about themselves and recording conditions through an anonymous registration form.
Specifically, we asked for age (a numerical integer), gender (male, female, non-binary, undeclared), region of origin in Italy, country of origin (if not Italian), instruction level, and presence of any speech impairment (e.g., lisp or stuttering). 
Recording conditions include the input device adopted (laptop, smartphone, or headphones) and environmental noise level (no noise to very noisy).
Such rich additional metadata will enable future per-group analyses, e.g., studying service quality of IC models over Italian regions. 

To ensure high-quality annotations, at least two individuals reviewed each sample. We consider a sample valid if 1) the utterance is intelligible from the recording and 2) it is coherent with the provided prompt. We validated the entire set of recordings as follows.
Once all samples were annotated, we ran a first validation round considering the entire collection. Then, we extracted all non-valid samples and ran a second annotation and validation round considering only those samples. We repeated the process until no non-valid samples were left.  

\subsection{Data Characterization}

\begin{figure}[!t]
  \centering
  \includegraphics[width=\linewidth]{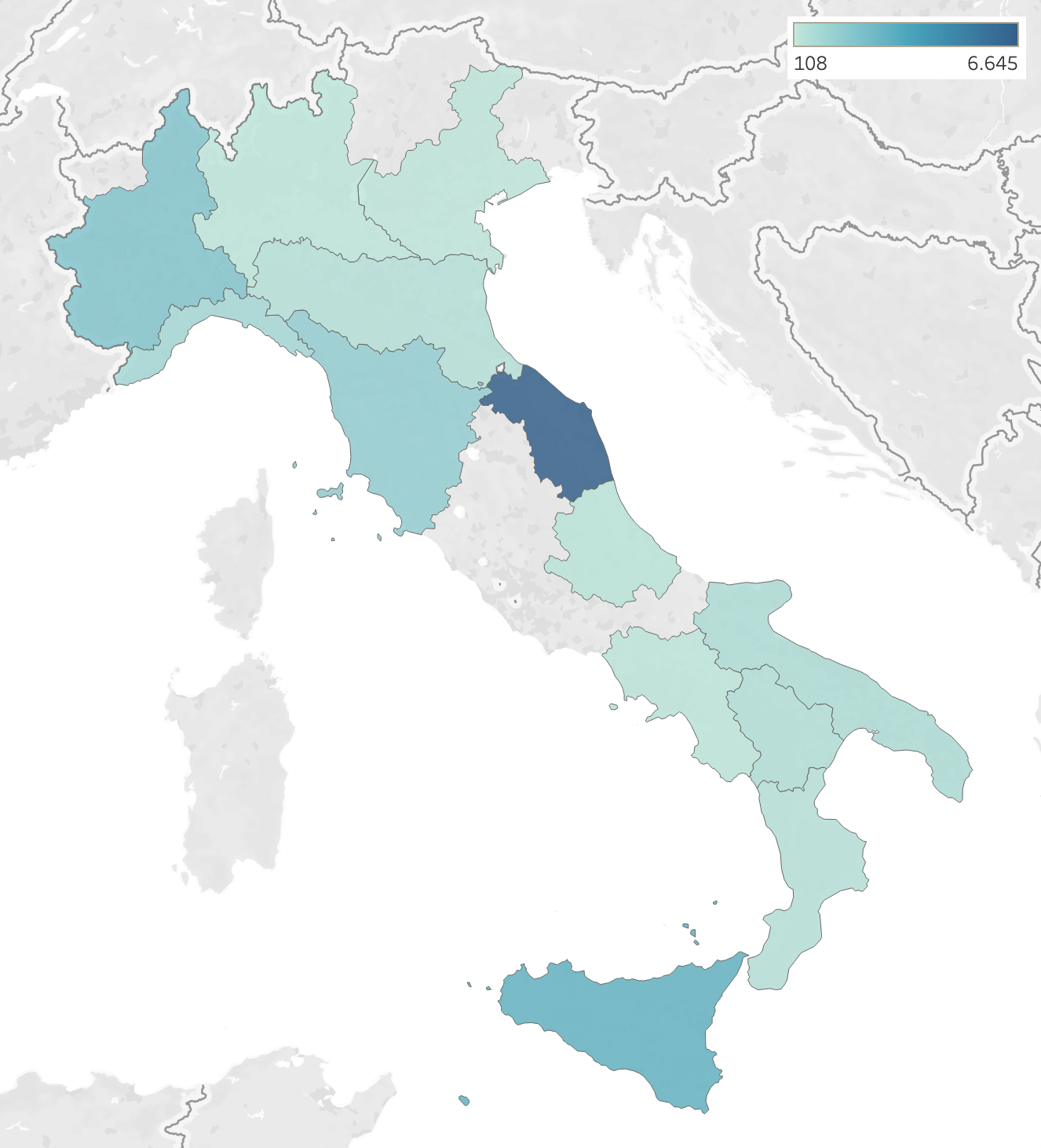}
  \caption{Distribution of utterances across Italian regions.
  Darker colors represent higher absolute counts.}
  \label{fig:region-distribution}
\end{figure}

We extracted and annotated every sample in the Italian split of MASSIVE~\cite{massive_dataset}, for a total of 15.46 hours of recording and 60 different intents.
The final dataset consists of 16,521 audio recordings by 70 distinct volunteers. All but one self-identified as native Italian speakers. 
Only four speakers declared some speech impairment.
Native speakers are distributed across 13 Italian regions, which concentrate a large number of linguistic and diatopic variations~\cite{italian_dialects_variations}.
Table~\ref{tab-data-demo} reports the distribution of gender and age in the dataset, while  Figure~\ref{fig:region-distribution} shows the distribution over the region of origin.

The audio length goes from 1.14 to a maximum of 38.34 seconds and an average of 3.37 seconds. We WAV-encoded recordings with a sampling frequency of 16 kHz.


 


\vspace{2mm}
\noindent \textbf{Dataset splits.} For experimental and reproducibility purposes, we release three splitting configurations of increasing difficulty\footnote{\label{rita_github}The dataset, splits, and annotation scheme are publicly available at \texttt{\repo}.}:

\begin{itemize}
    \item \textit{massive}: It uses the official training and test splits of MASSIVE \cite{massive_dataset} and includes all our speakers. Furthermore, the distribution of noisy utterances is randomly apportioned between the two partitions.
    \item \textit{speaker}: We stratify on speakers, i.e., all recordings of a speaker belong to either the training, validation, or test split.
    This division will help to assess whether the models generalize to unseen speakers;
    \item \textit{noisy}: The test set consists exclusively of data annotated with the highest background noise level, while the train and validation sets contain either noiseless or data with a low noise level, randomly split.
\end{itemize}

Table~\ref{tab:split_stats} reports statistics on the number of utterances, speakers, and hours of recordings for each splitting configuration. 

\begin{table}[!t]
    \caption{Dataset statistics.}
    \label{tab:split_stats}
    \centering
    \begin{tabular}{@{}llllccc@{}}
    \toprule
    \multicolumn{2}{l}{Configuration} & \# Utterances & \# Hours & \# Speakers\\ \midrule
    \multirow{3}{*}{massive} & train & 11514 & 10.80 & 69 \\
    & validation & 2033 & 1.90 & 68 \\
    & test & 2974 & 2.76 & 69 \\ \midrule 
    \multirow{3}{*}{speaker} & train & 13123  & 12.33 & 56 \\
    & validation & 1957 & 1.67 & 7 \\
    & test & 1441 & 1.46 & 7 \\ \midrule
    \multirow{3}{*}{noisy} & train & 13742 & 12.93  & 69\\
    & validation & 1526 & 1.44 & 66 \\
    & test & 1253 & 1.09 & 9 \\ \bottomrule
    \end{tabular}
\end{table}

\subsection{Tasks}

\textsc{ITALIC} enables a range of SLU and Natural Language Understanding (NLU) tasks.
In this paper, following the original setup in MASSIVE~\cite{massive_dataset}, we provide baselines for the intent classification and automatic speech recognition tasks.

Given the rich set of annotations we release, practitioners may use \textsc{ITALIC} in other tasks, such as speaker identification, text-to-speech, age estimation, or region of origin identification.
We leave the analysis in such contexts to future work.



\section{Experiments}


In this section, we present the experimental setup and results for evaluating the performance of various models on the \textsc{ITALIC} dataset. We focus on intent classification and automatic speech recognition. 
Our goal is to leverage the rich annotations in \textsc{ITALIC} to evaluate state-of-the-art models in terms of accuracy, robustness to noise, and generalization to unseen speakers.  
Additionally, this investigation seeks to explore the influence of two key factors, namely the knowledge of the Italian language and the variability of recording conditions, including variations in noise levels and speaker characteristics.

\subsection{Experimental Setting}

\textbf{Models.} We use established E2E transformer-based models for both IC and ASR setups.
We address IC using either raw audio signals or text transcripts.
As speech SLU models, we test several XLSR variants, a speech representation network pretrained on 53 \cite{xlsr_53} or 128 \cite{xlsr_128} languages. We conduct experiments on two model sizes, i.e., 300M and 1B parameters, testing XLSR-53 300M and XLSR-128 300M and 1B.
We build two additional baselines by testing
language adaptation on out-of-domain data. 
In practice, we employ two models, i.e., XLSR-53 300M and XLSR-128 1B, that have been fine-tuned via ASR on the Italian split of the Mozilla Common Voice dataset \cite{common_voice_ds}.
As text NLU models, we include multilingual BERT \cite{devlin-etal-2019-bert}, multilingual BART~\cite{mbart}, and two variants of these models pre-trained on Italian data only \cite{bert_it, BARTIT}. Note that both models include Italian as a pretraining language.

For the ASR task, we use Whisper \cite{whisper} in three variants: small (244M parameters), medium (769M), and large (1.5B). 

We fine-tune each IC model attaching to the encoder architecture a final classification layer.\footnote{In our experiments on intent classification, we use a classification layer on top of the BART encoder.}
We use model implementation and weights from the transformers library \cite{transformers_lib} in all experiments.

\vspace{2mm}
\noindent \textbf{Data.} We fine-tune models on the training split of one of the configurations of \textsc{ITALIC}, i.e., \textit{massive}, \textit{speaker}, or \textit{noisy}, and report the result on the test set.   

\vspace{2mm}
\noindent \textbf{Evaluation Metrics.} We measure accuracy and F1 Macro scores for IC and the Word Error Rate (WER) and Character Error Rate (CER) for ASR.

\vspace{2mm}
\noindent \textbf{Hyperparameter Setup.}
We ran a manual hyperparameter search and followed fine-tuning procedures and guidelines from relevant literature.


We provide detailed information about the models used for the evaluation, the hyperparameter setup, and the fine-tuning procedure in the official project repository.\footref{rita_github}

\subsection{Results on Intent Classification}


\begin{table}[!t]
    \caption{Accuracy and F1 Macro results of E2E-SLU models and adapted variants (FT: \checkmark). Best result per splitting configuration in bold.}
    \label{tab:e2e-slu_results}
    \centering
    \begin{tabular}{@{}llllll@{}}
    \toprule
    Split & Model & \# params & FT & Accuracy & F1 \\ \midrule
    \multirow{4}{*}{massive} & XLSR-128 & 300M &  & 76.16 & 76.11 \\
     & XLSR-128 & 1B &  & 77.07 & 77.08 \\
     & XLSR-53 & 300M & \checkmark & 81.34 & 81.31 \\
     & XLSR-128 & 1B & \checkmark & \textbf{83.39} & \textbf{83.25} \\ \midrule
    \multirow{4}{*}{speaker} & XLSR-128 & 300M &  & 73.42 & 73.04 \\
     & XLSR-128 & 1B &  & 79.11 & 79.08 \\
     & XLSR-53 & 300M & \checkmark & 83.69 & 83.62 \\
     & XLSR-128 & 1B & \checkmark & \textbf{84.18} & \textbf{84.05} \\ \midrule
    \multirow{4}{*}{noisy} & XLSR-128 & 300M &  & 78.29 & 78.21 \\
     & XLSR-128 & 1B &  & 76.48 & 76.06 \\
     & XLSR-53 & 300M & \checkmark & 81.01 & 80.94 \\
     & XLSR-128 & 1B & \checkmark & \textbf{82.20} & \textbf{82.43} \\ \bottomrule
    \end{tabular}
\end{table}

\vspace{2mm}
\textbf{E2E SLU.}
Table~\ref{tab:e2e-slu_results} reports the result of speech SLU models on the \textsc{ITALIC} dataset.
As expected, adaptation to Italian via ASR fine-tuning yields better models, with the adapted XLSR-128 1B model being the best across all splitting configurations.
Relative to the non-adapted version, XLSR-128 1B achieved +6.17 and +6.37 F1 points on the \textit{massive} and \textit{noisy} configurations, respectively. 
Except for one case (XLSR-128, \textit{noisy}), the results also suggest that larger models yield better results, although not as much as adapting models to Italian.

Notably, all models, except XLSR-128 300M, achieved higher performance on the challenging \textit{speaker} configuration compared to \textit{massive}, indicating their generalizability to different speakers and their ability to handle variations in speaking styles and accents. 
As expected, all models showed lower performance on the \textit{noisy} subset, with the only exception of the XLSR-128 300M model.
This finding highlights the impact of recording conditions on model performance, motivating the need for more resources and training procedures that closely match real-world scenarios.




\begin{table}[!t]
    \caption{Accuracy and F1 Macro results of text NLU models with multilingual (PT: M) and monolingual (PT: I) pretraining for the massive configuration. Best result in bold.}
    \label{tab:nlu_results}
    \centering
    \begin{tabular}{@{}lllll@{}}
    \toprule
    Model & \# params & PT & Accuracy & F1 \\ \midrule
    BART & 611M & M & 87.16 & 83.53\\
    BERT & 167M & M & 86.21 & 82.93 \\
    BART & 141M & I & 86.65 & 83.82  \\
    BERT & 110M & I & \textbf{88.43} & \textbf{85.57} \\ \bottomrule
    \end{tabular}
\end{table}

\vspace{2mm}
\noindent \textbf{NLU.} 
Table~\ref{tab:nlu_results} reports the results of addressing IC with text NLU models on the \textit{massive} splitting configuration. We do not test text models on \textit{speaker} and \textit{noisy} as these configurations are specifically customized for speech models and tasks.


Of particular interest is the performance of the Italian pre-trained BERT model, which, despite having fewer parameters, exhibits superior performance compared to the BART and BERT models pre-trained on multilingual data. Utilizing Italian data can yield valuable improvements in the performance of NLU models, even though the performance gap between the models is less pronounced. 
Our study demonstrates that, particularly for encoder-based models such as BERT, smaller models pre-trained on Italian data can achieve comparable or even superior performance compared to their larger multilingual counterparts.

Since two human inspectors have validated all \textsc{ITALIC} samples, we can safely assume that voice recordings closely match the text transcripts. We can then impute any difference in performance across modalities, i.e., speech and text, to the inherent difficulty of learning from the two kinds of raw data.
Interestingly, we note that the best text model, monolingual BERT 110M, achieves +2.32 F1 points, marking a gap despite having x9 fewer parameters. 
We can draw similar conclusions comparing other pairs of speech and text models.
These findings underscore the importance of a dedicated dataset for SLU tasks for better interpretation of the spoken language.




\subsection{Results on Automatic Speech Recognition}

\begin{table}[!t]
    \caption{WER and CER results of Whisper models in a zero-shot setup (S: ZS) and adapted variants (S: FT). Best result per splitting configuration in bold.}
    \label{tab:asr_results}
    \centering
    \begin{tabular}{@{}llllll@{}}
    \toprule
    Split & Model & \# params & S & WER & CER \\ \midrule
    \multirow{4}{*}{massive} & large & 1.5B & ZS & 11.46 & 5.01 \\
     & small & 244M & FT & 4.82 & 1.49 \\
     & medium & 769M & FT & 3.41 & 0.92 \\
     & large & 1.5B & FT & \textbf{3.06} & \textbf{0.82} \\ \midrule
    \multirow{4}{*}{speaker} & large & 1.5B & ZS & 8.65 & 3.93 \\
     & small & 244M & FT & 3.81 & 0.99 \\
     & medium & 769M & FT & 2.92 & 0.70 \\
     & large & 1.5B & FT & \textbf{2.74} & \textbf{0.61} \\ \midrule
    \multirow{4}{*}{noisy} & large & 1.5B & ZS & 15.41 & 7.67 \\
     & small & 244M & FT & 8.46 & 2.95 \\
     & medium & 769M & FT & 5.83 & 1.92 \\
     & large & 1.5B & FT & \textbf{5.29} & \textbf{1.70} \\ \bottomrule
    \end{tabular}
\end{table}

Although the \textsc{ITALIC} dataset was not specifically designed for ASR, its various speaker and recording conditions make it a valuable resource for analyzing the performance of Italian ASR models. 
For this task, we use the large Whisper model~\cite{whisper} in zero-shot settings and its fine-tuned version for Italian ASR in three different sizes: small, medium, and large. The results are presented in Table~\ref{tab:asr_results}. 

Our investigation reveals that all the evaluated models performed well, with low WER and CER scores (the estimated human WER is approximately 4\%~\cite{wer_human_4}).
Whisper large is best across the board, further highlighting the importance of model size.
As expected, there is a clear performance degradation on the \textit{noisy} configuration, more marked for smaller models. 

Applying Whisper to zero-shot speech recognition yields the worst performance on all splits, with a gap from the fine-tuned variant (Whisper large, S: FT) largest in \textit{noisy} and smallest in \textit{speaker}.
In absolute terms, it achieves a WER of 8.65 on the \textit{speaker} configuration and a much worse 15.41 on \textit{noisy}. These results are sensibly worse than standard Italian benchmarks such as Mozilla CV~\cite{common_voice_ds} (WER: 7.1) or Google Fleur~\cite{fleurs_ds} (WER: 4.0).

These findings prove \textsc{ITALIC} challenging for current state-of-the-art SLU and NLU models and underscore its importance as a novel Italian resource. With accurate recordings of real-world human-to-voice assistant interactions and rich annotations, \textsc{ITALIC} paves the way for new research and development of Italian models.

\section{Conclusions}

We presented \textsc{ITALIC}, the first large-scale Italian audio dataset specifically designed for intent classification. The collection is comprehensive of text transcripts, recordings, and additional metadata about the speaker and the recording channel.
We evaluated the performance of current state-of-the-art speech and text models on the intent classification and automatic speech recognition tasks, demonstrating the impact of model selection and pretraining on performance. 
We release the dataset, annotation schema, and code to foster future research in this area.

Our future work includes expanding the \textsc{ITALIC} dataset to enhance its diversity and representativeness of the Italian language and exploring new tasks that can be tackled with this dataset. 
Future enhancements will involve adding more speakers with diverse backgrounds, including non-native speakers, and further extending the dataset to address any potential gap in coverage.
We also aim to develop a large-scale multilingual data collection platform to facilitate the creation of similar datasets in other languages.

\section{Limitations}


The \textsc{ITALIC} dataset is valuable for evaluating models for the Italian SLU and ASR tasks.
However, some limitations must be taken into account when interpreting the results.
While the dataset includes recordings from a wide range of Italian regions, it only partially represents all dialects and linguistic varieties. 
Additionally, the dataset is mainly composed of recordings from native Italian speakers, which may not be representative of scenarios where the user of a voice assistant has a non-native accent.
We envisioned the dataset to represent a broad spectrum of individuals, from non-binary to speakers with speech impairments—however, only a limited number of volunteers identified as such. We will promote future dataset releases capturing more speech nuances.
Finally, the dataset only includes one recording per sentence. 
Including multiple recordings of the same sentence by different speakers would allow a more comprehensive evaluation of model performance, which is of key importance for SLU domain~\cite{koudounas2023}.

Overall, the \textsc{ITALIC} dataset provides a strong foundation for evaluating Italian SLU and ASR models; addressing these limitations will enable more comprehensive evaluations and further advances in these fields.

\section{Acknowledgments}
This project is a joint effort of members of the ``Risorse per la Lingua Italiana'' open community. We would like to thank all the crowd workers who participated to our campaign and the reviewers for their helpful comments. 
This work is partially supported by the FAIR - Future Artificial Intelligence Research (PE00000013)  and the spoke ``FutureHPC \& BigData'' of the ICSC - ``National Centre for HPC, Big Data and Quantum Computing'' funded by the European Union Next-GenerationEU, and SmartData@PoliTO center on Big Data and Data Science.
This manuscript reflects only the authors views and opinions, neither the European Union nor the European Commission can be considered responsible for them.

\bibliographystyle{IEEEtran}
\bibliography{mybib,anthology}

\end{document}